\documentclass{article}

\PassOptionsToPackage{numbers, compress}{natbib}
\usepackage[final]{neurips_2023}

\usepackage[utf8]{inputenc} % allow utf-8 input
\usepackage[T1]{fontenc}    % use 8-bit T1 fonts
\usepackage{hyperref}       % hyperlinks
\usepackage{url}            % simple URL typesetting
\usepackage{booktabs}       % professional-quality tables
\usepackage{amsfonts}       % blackboard math symbols
\usepackage{nicefrac}       % compact symbols for 1/2, etc.
\usepackage{microtype}      % microtypography
\usepackage{xcolor}         % colors
\usepackage{amsmath}
\usepackage{graphicx}
\newtheorem{theorem}{Theorem}
\usepackage{wrapfig}
\usepackage{array}
\usepackage{multirow}
\usepackage[toc]{appendix} 
\usepackage{algorithm}
\usepackage{algpseudocode}

\hypersetup{colorlinks,linkcolor={blue},citecolor={blue},urlcolor={blue}}
\title{DynaLay: An Introspective Approach to Dynamic Layer Selection for Deep Networks}

\author{%
  Mrinal Mathur\thanks{Correspondence to: \texttt{mmathur4@student.gsu.edu}, \texttt{s.m.plis@gmail.com}} \\ TReNDS Center \thanks{The Georgia State University/Georgia Institute of Technology/Emory University Center for Translational Research in Neuroimaging and Data Science (TReNDS Center).} , Georgia State University\\
  \And
  Sergey Plis \footnotemark[1] \\
  TReNDS Center \thanks{The Georgia State University/Georgia Institute of Technology/Emory University Center for Translational Research in Neuroimaging and Data Science (TReNDS Center).}, Georgia State University \\
}

\begin{document}

\maketitle

\begin{abstract}
  Deep learning models have become increasingly computationally intensive, requiring extensive computational resources and time for both training and inference. A significant contributing factor to this challenge is the uniform computational effort expended on each input example, regardless of its complexity. We introduce \textbf{DynaLay}, an alternative architecture that features a decision-making agent to adaptively select the most suitable layers for processing each input, thereby endowing the model with a remarkable level of introspection. DynaLay reevaluates more complex inputs during inference, adjusting the computational effort to optimize both performance and efficiency. The core of the system is a main model equipped with Fixed-Point Iterative (FPI) layers, capable of accurately approximating complex functions, paired with an agent that chooses these layers or a direct action based on the introspection of the models inner state. The model invests more time in processing harder examples, while minimal computation is required for easier ones. This introspective approach is a step toward developing deep learning models that "think" and "ponder", rather than "ballistically'' produce answers. Our experiments demonstrate that DynaLay achieves accuracy comparable to conventional deep models while significantly reducing computational demands.

\end{abstract}

\section{Introduction}
In the evolving realm of deep learning, the escalating complexity and
scale of models have driven breakthroughs in key areas such as natural
language processing, computer vision, and reinforcement learning
 \cite{schmidhuber2015deep, vaswani2017attention, Hochreiter1995LONGST}. However, this advancement is accompanied by soaring computational demands, posing significant challenges, particularly in scenarios with limited computational resources. A critical inefficiency in traditional deep learning models is their uniform computational expenditure across diverse inputs, regardless of each input's intrinsic complexity. This one-size-fits-all strategy results in suboptimal resource utilization, as not all inputs require the same level of computational effort. For instance, simple inputs that a model can easily interpret may not necessitate the full extent of the network’s processing capabilities, whereas more complex inputs might benefit from additional computation for accurate interpretation. This discrepancy highlights a need for a more dynamic and discerning approach in deep learning architectures. Such an approach would intelligently adapt the computational effort to the complexity of each input, ensuring that computational resources are judiciously allocated, optimizing efficiency without compromising on accuracy.

To tackle these challenges, we have developed a prototype architecture that employs an introspective mechanism, powered by a decision-making agent. This agent receives internal state of the model---its activations---as an input and dynamically selects the appropriate number of layers of the deep model to iterate over until fixed point for each specific input, thereby aligning computational efforts with how difficult the input is for the model. This introspective capability imbues the model with an exceptional level of adaptability, enabling it to recalibrate its structure on-the-fly. This real-time optimization harmonizes the goals of maintaining high classification accuracy while enhancing computational efficiency. Central to our architecture are Fixed-Point Iterative (FPI) layers, celebrated for their proficiency in approximating intricate functions with a streamlined parameter set \cite{arya2022automatic, kidger2022neural}. The agent is meticulously crafted to judiciously toggle between these FPI layers or to choose a no-operation (NOP) stance, contingent on the nuanced demands of each input. To further refine this adaptive decision-making process, we introduce a novel reward function. This function is designed to incentivize the agent to deeply analyze and learn from the main model, fostering an introspective, context-sensitive approach to layer selection and computational allocation.

Our principal contributions through the development of \textbf{DynaLay} are multifaceted and innovative:

\begin{enumerate}
\item Introspective Architecture: We introduce an advanced
  architecture that synergizes a deep learning model, equipped with
  Fixed Point Iterative (FPI) layers, and an auxiliary agent
  network. This unique combination enables dynamic selection of the
  FPI layers, tailoring the model's computational efforts to each
  input's complexity as reflected by internal model's
  state \cite{arya2022automatic, kidger2022neural}.
\item Observer Model Integration: A pivotal feature of our architecture is the incorporation of an observer model. This model vigilantly monitors the internal dynamics of the predictive model. In instances where the predictor seems likely to err, as evidenced by less-than-ideal internal representations, the observer intervenes. It compels the predictor to iterate further, achieving convergence. This mechanism allows for a dynamic adjustment in computational expenditure, allocating more resources for complex inputs and conserving energy on simpler ones 
 \cite{rumelhart1986learning, schmidhuber2015deep}.
\end{enumerate}

The structure of the manuscript is methodically arranged to facilitate a comprehensive understanding of our work: Section 2 lays the foundational knowledge about FPI layers and the role of reinforcement learning in deep learning contexts. Section 3 extensively explores the intricacies of our proposed architecture, elucidating its unique components and operational principles. Section 4 is dedicated to our experimental approach and the presentation of results, showcasing the effectiveness of DynaLay. Finally, Section 5 wraps up the manuscript with a summary of our findings, reflections on the significance of DynaLay, and perspectives on future research directions in this exciting field \cite{vaswani2017attention, Hochreiter1995LONGST}.

\begin{figure}
    \centering
    \includegraphics[width=\textwidth]{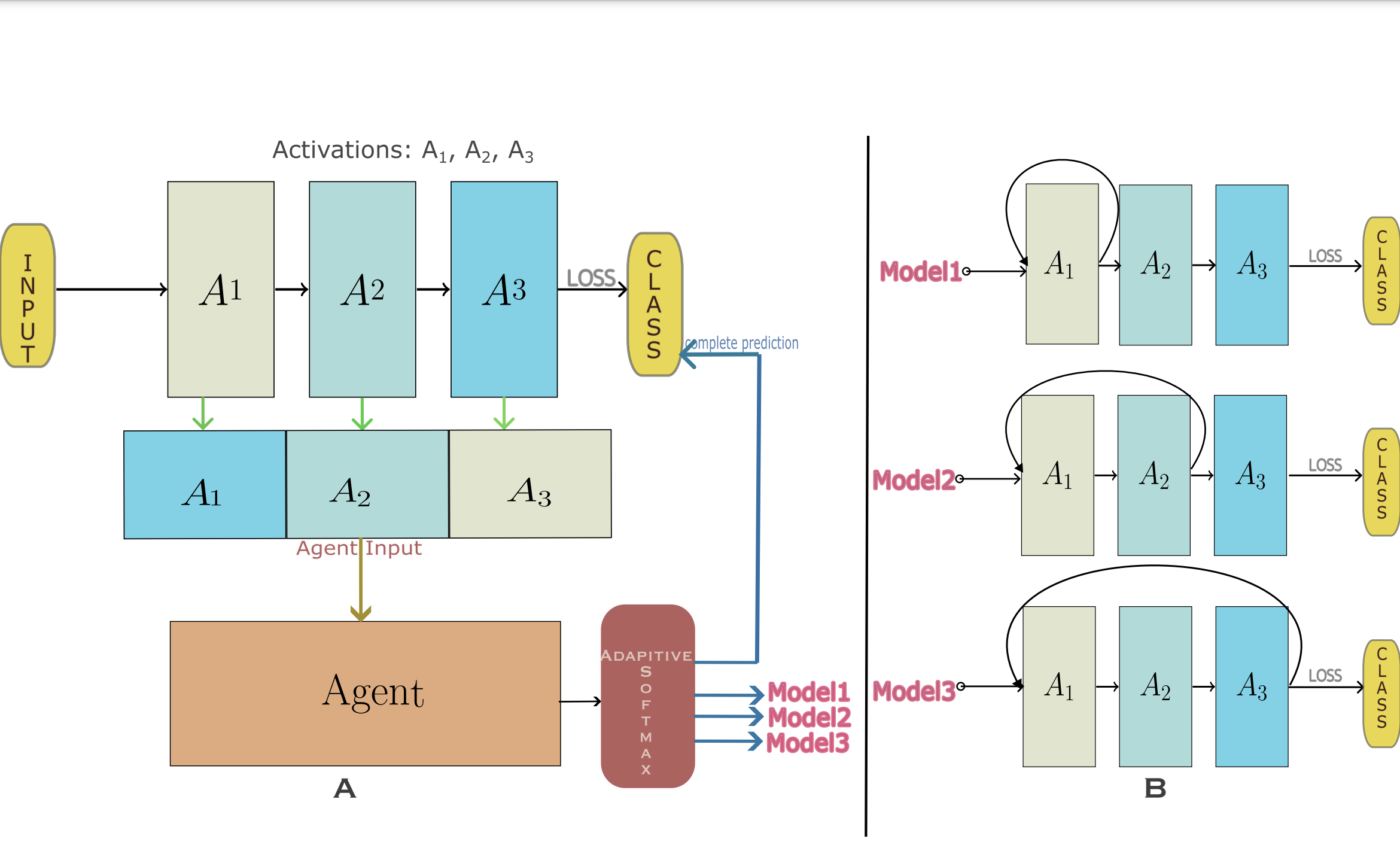}
    \caption{Main Model Architecture: A$)$ Schematic representation of the DynaLay architecture, highlighting its key components and the flow of data.B$)$ The figure illustrates how the agent dynamically selects layers for Fixed-Point Iteration, where $A_i$ for the same $i$ are parameter tied \cite{chatziafratis2022scrambling, habiba2021neural}}
    \label{fig:architecture}
\end{figure}

\section{Background}

\subsection{Fixed-Point Iterative Layers in Deep Learning}

Fixed-point iterative (FPI) methods have become increasingly significant in deep learning, offering a robust approach to handling complex, non-linear equations. These methods revolve around the concept of finding a fixed point, a point that remains constant under a specific function, which is pivotal in numerous deep learning applications. Fixed-point iterative (FPI) \cite{bolte2022automatic, werbos1990backpropagation} methods are a class of techniques commonly used to solve equations of the form $ f(x) = x $, where $ x $ is a fixed point  \cite{habiba2021neural} of the function $ f $. In deep learning, FPI layers are designed to find a fixed point $ z $ that satisfies:$
z = f(z, x)
$. Here, $ f $ is the neural network layer's function, $ x $ is the input to the layer, and $ z $ is the fixed-point that we want to compute. The general iterative scheme for finding this fixed point can be expressed as:
$
z^{(k+1)} = f(z^{(k)}, x)
$
where $ z^{(k)} $ is the estimate at iteration $ k $. The process begins with an initial guess $ z^{(0)} $ and iterates until convergence, i.e., $ || z^{(k+1)} - z^{(k)} || < \epsilon $ where $ \epsilon $ is a small tolerance.

we also define $\hat{x}$ which is the fixed point of $g$, which satisfies $\hat{x} = g(\hat{x})$, where $\hat{x}$ is the converged vector of $g$ function. Few important concepts to define fixed point iterations involves Contrastive Mapping \cite{bolukbasi2017adaptive, mackay2018reversible}.

\begin{theorem}[Contraction mappings for fixed point operation]

Consider a function $ f: \mathcal{X} \to \mathcal{X} $ where $ \mathcal{X} $ is a complete metric space. Let $ g(x, \theta, l) $ be the output when using layer $ l $ in our neural network with input $ x $ and parameters $ \theta $. Assume that $ f $ is a contraction mapping on $ \mathcal{X} $, i.e., there exists a $ \beta \in [0, 1) $ such that:
$
d(f(x), f(y)) \leq \beta \cdot d(x, y), \quad \forall x, y \in \mathcal{X}
$

where $ d(\cdot, \cdot) $ is a distance metric on $ \mathcal{X} $.

Under these conditions, our adaptive layer selection mechanism 
 \cite{jeon2021differentiable, rajeswaran2019meta} that minimizes $ ||f(x, \theta) - g(x, \theta, l)||_2^2 $ will converge to a unique fixed point $ x^* $ in $ \mathcal{X} $, such that:
$$
x^* = f(x^*)
$$

It is especially  valuable since we relate $ \beta $ to the chosen layer, which makes the contraction faster based on layer attributes. 
\label{theorum:contraction}
\end{theorem}

One of the unique features of our approach in \textbf{DynaLay} is the deliberate avoidance of explicit fixed-point solvers , such as those based on the Banach fixed-point theorem \cite{agarwal2018banach}, commonly employed in methods like Deep Equilibrium Models (DEQs) \cite{bai2019deep}. Iterative solvers can introduce numerical instabilities, particularly for ill-conditioned problems. Our architecture, by adhering to contraction mapping principles as mentioned in Theorem \ref{theorum:contraction}, guarantees stability. Formally, if $ T: X \rightarrow X $ is a contraction mapping, then for every $ x, y \in X $, 
$
d(T(x), T(y)) \leq k \times d(x, y)
$
where $ 0 < k < 1 $, ensuring stability.
The absence of a solver simplifies both the architecture and the training algorithm, making the model easier to implement and tune.
Our method is rooted in the principles of contraction mapping, providing a robust theoretical foundation that guarantees both the existence and uniqueness of a fixed-point solution without necessitating an explicit solver as mentioned in Equation \ref{ref:solver}: 
\begin{equation}
\text{If } T: X \rightarrow X \text{ is a contraction, then } \exists ! x^* \in X : T(x^*) = x^*    
\label{ref:solver}
\end{equation}

Our layer selection mechanism is designed to work well with Theorem  \ref{theorum:contraction} adhering to its principles to ensure the model's stability and convergence. This obviates the need for a solver while still offering robust theoretical guarantees as explained in next section.

\subsection{Layer Selection Mechanism}
In traditional architectures and equilibrium networks \cite{bai2019deep}, the network essentially learns a single fixed point. It may not always adapt well to the diverse computational needs of different inputs. 
 \cite{bai2019deep} assumes that all layers have the same computational requirements, which may not always be the case.

On the other hand, \cite{banino2021pondernet} introduces a more dynamic approach to network architecture by adding or pruning neurons or layers based on the problem at hand. While this approach offers a more flexible model, it comes with its own set of challenges. The most significant challenge is the risk of making the architecture unstable using ACT \cite{graves2016adaptive}, which in turn can lead to decreased performance in actual predictions. Additionally, the process of modifying the architecture may be computationally expensive.

We define the computational cost of our model as the sum of time take for each layer computation after applying fixed point iteration on it as  $C(n, t) = \sum_{i=1}^{n} t_i$,
where 
$ n $: Number of layers in the neural network, $ t_i $: Time taken for computation at the $ i^{th} $ layer and  $ C(n, t) $: Computational cost of our model, defined as the sum of time taken for each layer computation.
Our model introduces a layer-selection mechanism to achieve the aforementioned objective where, for each input, the model employs an agent that uses the current activation's and the inherent characteristics of the data to selectively choose a subset of layers $S$ from the available layers $n$. this can be represented as: $C(S, t) \leq C(n, t) < C_0$.  The computational graph then proceeds only through the layers in $S$, thereby avoiding the computational cost of the unselected layers.

In this paper, we empirically validate the computational efficiency of our model by measuring $ C(n, t) $ across multiple data points. Our results indicate that, on average, $ C(n, t) $ is consistently less than $ C_0 $, thereby confirming the computational advantages of our approach as shown in \ref{fig:impact_of_fpi}

Our approach aims to combine the best of both worlds. We introduce an agent-based mechanism for layer selection that dynamically decides which layers of the network to use for each input sample as seen in Figure \ref{fig:architecture}. This allows the network to adapt its complexity based on the specific computational needs of each input, without the need to alter the underlying architecture. As a result, our method maintains the stability of the architecture while offering a tailored approach to problem-solving. 

\section{Methods}

\subsection{Model Architecture}

In this paper, we introduce a groundbreaking architecture, \textbf{DynaLay} (Dynamic Layer Selection), which is elucidated in Figure \ref{fig:architecture}. Our architecture injects a layer of introspection into neural networks by employing an adaptive layer-selection mechanism. Specifically, an agent \( A \) takes input activations and makes an introspective decision to select the most appropriate layer or set of layers \( l \) for each sample \( x^{(i)} \) in a batch. This forms a composite function \( g: (x^{(i)}, a) \rightarrow y^{(i)} \), where \( g \) represents the neural network with dynamically chosen layers based on the unique characteristics of each input sample.

The agent \( A \) is trained using a novel loss function that encapsulates classification accuracy \cite{schmidhuber2015deep} and computational cost within a reward function \( R \). This introspective capability allows the network to optimize the trade-off between performance and computational efficiency, dynamically adjusting the network's complexity according to the individual characteristics of each sample \( x^{(i)} \).

While many established architectures \cite{mackay2018reversible, vaswani2017attention, Hochreiter1995LONGST, he2016deep, marblestone2020product} utilize mechanisms such as skip connections or attention to optimize performance, these techniques are often task-dependent and do not adapt to individual samples within a dataset. \textbf{DynaLay} diverges from this approach. \textbf{Our core innovation lies in the model's introspective ability to recompute the results on more difficult inputs during inference and adaptively select layers for each sample based on its unique activation profile, optimizing both accuracy and computational efficiency}. Unlike previous work such as Adaptive Computation Time (ACT) \cite{graves2016adaptive}, which focuses on adjusting the number of recurrent steps for each sample in a sequence model, \textbf{DynaLay} offers finer granularity in layer-wise adaptivity and considers the nuanced contributions of different layers to the final model output for each individual sample.

In contrast, \textbf{DynaLay} not only allows for dynamic layer selection in both sequence and feed-forward architectures like CNNs 
 \cite{memory2010long}, but also provides a more nuanced approach by considering the activation profiles of each layer for every sample. 

Let $ L = \{l_1, l_2, ..., l_n\} $ be the set of layers in the \textbf{Main} Network, and $ A_i $ be the activation output from layer $ l_i $. The \textbf{Agent} Network is designed to take activation $ \{A_1, A_2, ..., A_n\} $ as input and output a probability distribution $ P = [p_1, p_2, ..., p_n] $ defined as $ p_i = \text{AgentNet}(A_i) $, where $ \sum_{i=1}^{n} p_i = 1 $. A layer $ l $ is adaptively chosen based on this probability distribution $ P $, formalized as $ l = \text{Sample}(P) $. This is designed in such a way that the smaller (agent) model will not affect the distribution of larger model. Since we want our agent to maximize accuracte decision, we need to introduce a reward function that can guide our model to accurate layers.

Our reward function $ R $ is defined as 
\begin{equation}
 R = \alpha \times \text{Classification Accuracy} - \beta \times \text{Computational Cost} \times \text{Consecutive Layer Count}
\end{equation} with $ \alpha $ and $ \beta $ as hyperparameters. This guides the Agent Network's decision-making process to optimize both classification accuracy and computational efficiency. This reward function also counts the penalty by calculating the amount of time our model takes on a particular layer which is given by consecutive layer count and computation cost. Since the reward function and penalty are the main elements for the working of our agent.

\begin{equation}
    \mathcal{L} = \text{Cross-Entropy Loss} - \gamma \times R 
    \label{ref:loss}
\end{equation}

We also define a custom loss function $ \mathcal{L} $ for the entire architecture incorporates this reward function as Equation \ref{ref:loss}, where $ \gamma $ is a hyperparameter.

This architecture allows \textbf{DynaLay} to not only provide dynamic layer selection in both sequence and feed-forward architectures, but also offers a more nuanced approach by considering the activation profiles of each layer for every sample. This results in a more effective and computationally efficient model that adapts to the complexities of individual samples across a wide range of tasks.

\subsection{Backward Pass}
One of the key challenges in training \textbf{DynaLay} is the efficient computation of gradients during the backward pass. Given our architecture's dynamic nature and the presence of FPI layers, traditional backpropagation methods prove to be suboptimal 
 \cite{chung2016hierarchical}. The gradient flow can be tricky if there are multiple networks involved especially in teacher student and reinforcement learning architectures \cite{tessera2021keep, liu2018darts}. The gradient flows through the main architectures through all the layers in the first pass and then flows through the agent class to make the deicision, during which gradient flow is stopped from the main network. This complex flow of gradient is very important to keep in check. Hence, we device few techniques on top of backpropogation which used is used to keep the gradients in check.

\subsubsection{Gradient Accumulation for Dynamic Layer Selection}

For each sample, we have a unique set of layers $ L_s = \{l_1, l_2, ..., l_k\} $ chosen by the Agent Network. During backpropagation, we accumulate the gradients for each chosen layer $ l $ using a modified gradient accumulation technique, described in \cite{arya2022automatic, bolte2021nonsmooth}  as specified in Equation \ref{ref:grad_accumulate} 

\begin{equation}
\frac{\partial \mathcal{L}}{\partial W_l} = \sum_{s \in S} \frac{\partial \mathcal{L}_s}{\partial W_l} \quad \text{for each } l \in L_s
\label{ref:grad_accumulate}
\end{equation}

where $ S $ is the set of all samples, $ \mathcal{L}_s $ is the loss for sample $ s $, and $ W_l $ are the parameters of layer $ l $.

\subsubsection{Optimized Backpropagation for FPI Layers}
For the FPI layers \cite{jeon2021differentiable}, we adapt the methodology proposed in \cite{bolte2021nonsmooth}. This method allows us to compute the gradients implicitly, thereby reducing the computational complexity.
Let $( F(x) = 0$ represent the fixed-point equation for the FPI layer as shown in Equation \ref{ref:grad_opt}. Then the gradient $ \frac{\partial F}{\partial x} $ can be computed implicitly as:
\begin{equation}
\left( I - \frac{\partial F}{\partial x} \right)^{-1} \frac{\partial F}{\partial W} = - \frac{\partial F}{\partial W} 
\label{ref:grad_opt}
\end{equation}
where $ I $ is the identity matrix and $ W $ are the parameters of the FPI layer.
Typically, we would calculate $\frac{\partial F}{\partial x}$ during backpropagation \cite{rumelhart1986learning}. However, this approach can be highly inefficient and unstable. Instead, we use orthogonal over-parameterization \cite{arya2022automatic, metz2022gradients} to find a more effective representation as shown in Equation \ref{ref:grad_orth}:

\begin{equation}
\frac{\partial \mathcal{L}}{\partial W} = O \left( \frac{\partial \mathcal{L}}{\partial \tilde{W}} \right) 
\label{ref:grad_orth}
\end{equation}
where $ O $ is the orthogonal matrix and $ \tilde{W} $ is the over-parameterized representation of $ W $ \cite{metz2022gradients}

\subsection{Agent Network (\textbf{AgentNet} )}

\begin{figure}
    \centering
    \includegraphics[width=0.8\textwidth,height=50mm]{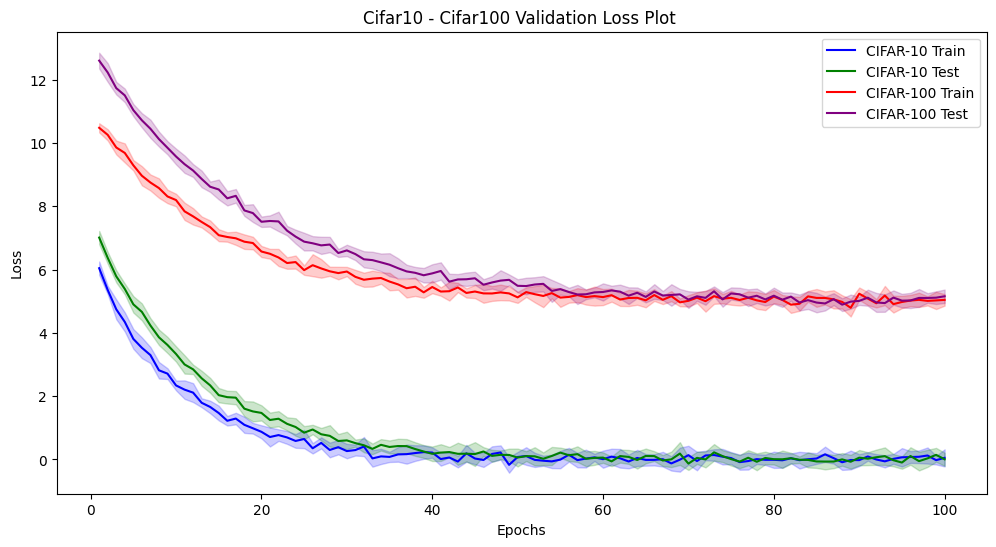}
    \caption{The plot illustrates the evolution of training and testing loss for both CIFAR-10 and CIFAR-100 datasets over 100 epochs. Each curve represents the mean loss across 10 different random seeds, with the shaded regions indicating one standard deviation from the mean}
    \label{fig:cifar10-100}
    
\end{figure}
The Agent Network serves as the linchpin in the \textbf{DynaLay} architecture, introducing an introspective dimension uncommon in traditional neural network designs. Unlike conventional architectures where each layer processes every sample in a uniform manner, our Agent Network exercises discernment in selecting the most relevant layers for individual samples. This adaptability optimizes both computational efficiency and model accuracy. Let $x$ represent the input to the architecture, and \( a_l \) denote the activation map for the \( l^{th} \) layer. These activation maps \cite{chen2023extracting} are scrutinized by the Agent Network to generate a probability distribution \( P \) over the layers. Mathematically, this distribution is expressed as: $$
P = \text{Adaptive Softmax} (F(a_1, a_2, \ldots, a_L))$$
Here, the function \( F \) integrates the activation maps \( a_1, a_2, \ldots, a_L \) from different layers, and the Adaptive Softmax function \cite{grave2017efficient} transforms these into probabilities, enabling the agent to make informed, adaptive decisions on layer selection for each specific sample.
Here, $ F $ is the function instantiated by the Agent Network, and $ L $ is the total number of layers in the architecture. The $ P_l $ component of $ P $ signifies the probability that the $ l^{th} $ layer should be activated for the specific input sample $ x $.

where $ \alpha $ and $ \beta $ are hyperparameters that manage the balance between classification accuracy ($ \text{Acc} $) and computational cost ($ C $) and Layer Count is which layer is chosen. During the training process, back propagation \cite{rumelhart1986learning} is employed to not only minimize the conventional loss $ \mathcal{L} $ but also to maximize the reward $ R $. The total loss function is thus formulated as: $$\text{Total Loss} = \mathcal{L} - \gamma \times R$$ where, $ \gamma $ is a hyperparameter that controls the significance of the reward term in the overall loss function.
\section{Experiments}
To validate the efficacy of our proposed architecture, \textbf{DynaLay}, we conducted experiments on multiple datasets, including image datasets:  CIFAR-10 \cite{giuste2020cifar} and CIFAR-100 \cite{krizhevsky2009learning}and text dataset: WikiText-2 
 \cite{merity2016pointer} and IMDB \cite{maas2011learning}. The primary goals of these experiments are twofold: 1) to evaluate the classification accuracy of our model in comparison with existing state-of-the-art architectures, and 2) to assess the computational efficiency improvements achieved through our adaptive layer selection mechanism.
We implemented all our models in PyTorch \cite{paszke2019pytorch} and conducted training on NVIDIA A40 GPUs with 40GB memory. For optimization, we chose the Adam optimizer \cite{kingma2014adam}, setting the learning rate at $1 \times 10^{-3}$ and the batch size at 64. Hyperparameters $ \alpha $, $ \beta $, and $ \gamma $ were optimized to 0.5, 0.3, and 0.2, respectively, through 9-fold cross-validation. All models were trained over 100 epochs. The tolerance for fixed-point iterations in the \textbf{DynaLay} architecture was set to $1 \times 10^{-5}$.

\subsection{Comparative Analysis}
In the context of adaptivity, DynaLay incorporates an AgentNet mechanism that enables dynamic layer selection, setting it apart from the more rigid 3-layer model and the solver-dependent adaptivity in \cite{bai2019deep} models. The AgentNet mechanism not only enhances adaptability but also contributes significantly to computational efficiency.

Table \ref{table:comparison} offers a comprehensive comparison between DynaLay, a standard 3-layer model, and DEQ \cite{bai2019deep} models equipped with solvers \cite{agarwal2018banach}. A salient feature of DynaLay is its remarkable efficiency in backward pass computations. Specifically, it outclasses both the standard 3-layer and DEQ \cite{bai2019deep} models in backward pass time for CIFAR10 and CIFAR100, clocking in at 87 and 90 minutes, respectively. This efficiency is pivotal for scaling and real-time applications.

Furthermore, DynaLay achieves competitive test accuracies of 92\% and 85\% on CIFAR10 and CIFAR100, respectively. This implies that the computational advantages of DynaLay do not compromise its performance, reinforcing its utility as a robust and efficient architecture.

\begin{table}[h]
\small
\centering
\caption{Comparison of Computational Steps for Different Models}
\label{table:comparison}
\begin{tabular}{|c|c|c|c|c|c|}
\hline
\multirow{2}{*}{\textbf{Aspect}} & \multicolumn{2}{c|}{\textbf{DynaLay}} & \multicolumn{2}{c|}{\textbf{3 layer CNN Model}} & \multicolumn{1}{c|}{\textbf{DEQ with Solvers}} \\
\cline{2-6}
& \textbf{CIFAR10} & \textbf{CIFAR100} & \textbf{CIFAR10} & \textbf{CIFAR100} & \textbf{CIFAR10}  \\
\hline
Backward Pass Time (secs) & \textbf{87} & \textbf{90} & 120 & 140 & 600(mins)  \\
\hline
Adaptivity & \multicolumn{2}{c|}{\textbf{AgentNet}} & \multicolumn{2}{c|}{None} & \multicolumn{1}{c|}{Solver-based} \\
\hline
Test Accuracy (\%) & \textbf{92} & \textbf{85} & 88 & 82 & 91  \\
\hline
\end{tabular}
\end{table}

\subsection{Dynamic Layer Selection in DynaLay}
\begin{figure}[h]
    \centering
    \vspace{-10pt}
    \includegraphics[width=0.8\textwidth,height=50mm]{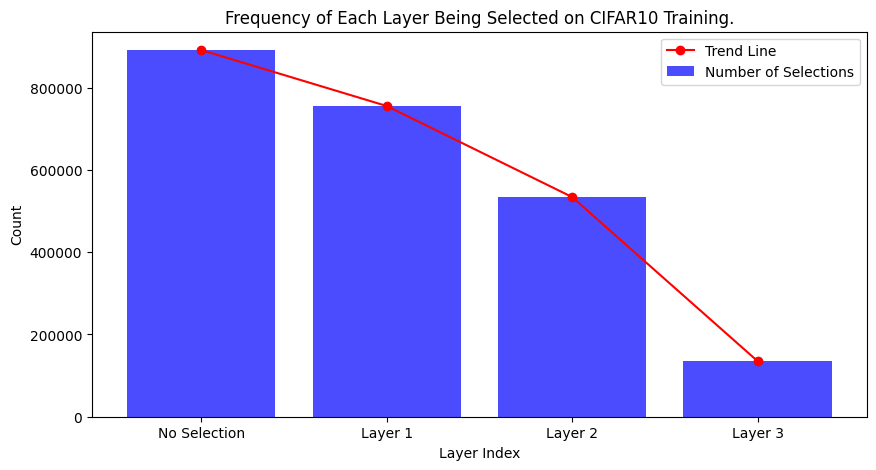}
    \caption{Frequency distribution of layer choices during CIFAR10 training. The bar chart quantifies the selection frequency of each layer across numerous training epochs, offering insights into their relative importance for CIFAR10 performance.}
    \label{fig:Layer_Selection}
\end{figure}

During our experiments on checking the frequency at which layers are selected, remarkably, the "No Layer" or Straightforward option emerges as the most frequently selected, indicating its ability to capture essential features across a broad spectrum of tasks and that the samples becomes easier once they are learned so the layers dont have to spend a lot of time to process them. Conversely, Fixed Point functionalities on Layer1 and Layer2 are invoked 25\% and 20\% of the time, respectively meaning that maybe initially the samples are difficult and takes more time to be processed by the network. The diminishing frequency of Layer3 selections implies that the model tends to minimize reliance on this layer's fixed-point operations as it stabilizes. 
Figure \ref{fig:Layer_Selection} elucidates the frequency distribution of each layer's selection during the evaluation process. 
%{r}{0.5\textwidth}

This dynamic layer selection mechanism substantially augments DynaLay's computational efficiency. By intelligently and adaptively 
 \cite{bolte2022automatic, jeon2021differentiable} deciding the appropriate layer for FPI \cite{rajeswaran2019meta} execution, DynaLay eliminates redundant computations, especially when less complex layers are sufficient for the task at hand \cite{bolukbasi2017adaptive}. This adaptability stands in sharp contrast to traditional architectures like CNNs and LSTMs 
 \cite{memory2010long}, which are constrained by their rigid, predetermined structures.

\subsection{Loss Convergence on CIFAR-10 and CIFAR-100}
The convergence behavior of our model's training and testing loss is a critical aspect of its evaluation. Figure \ref{fig:cifar10-100} provides a detailed insight into this aspect. The similarity in convergence patterns between CIFAR-10 and CIFAR-100 also suggests that our model is scalable and adaptable to different tasks and data distributions, aligning well with the core objectives of this research project. We also showed experiments on LSTM stacked layer in Appendix. 

\subsection{Influence of Pre-training on DynaLay's Efficacy}
\begin{figure}[h]
\centering
\includegraphics[width=\textwidth]{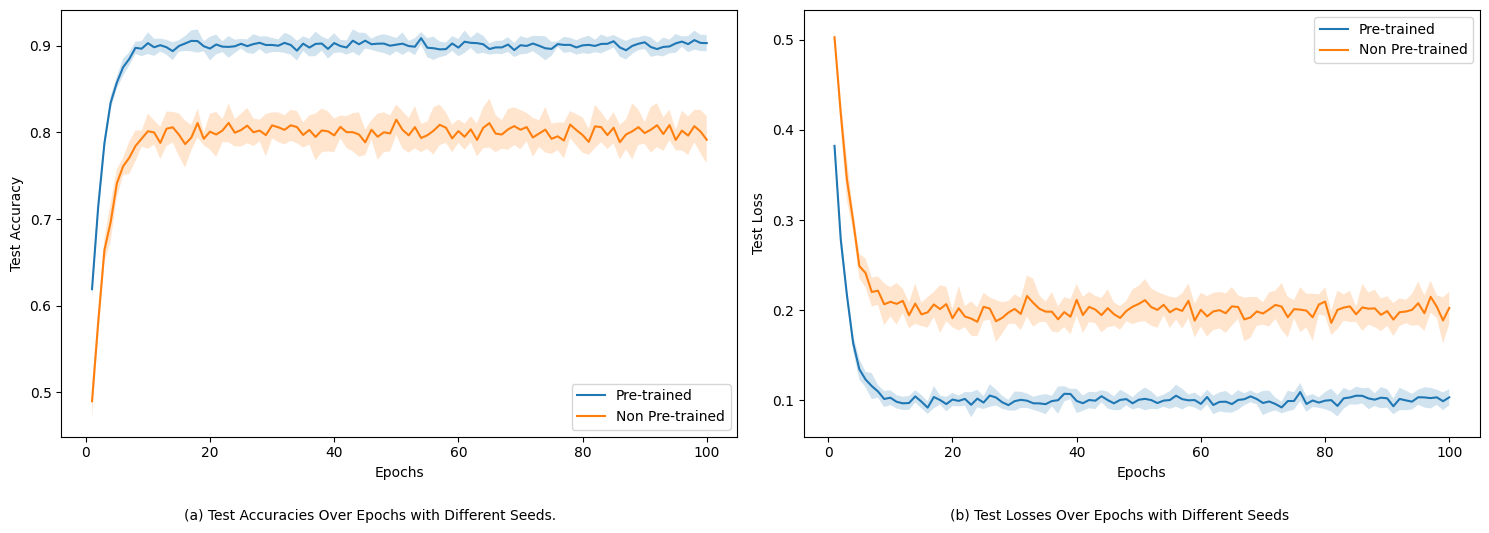}
\caption{Temporal evolution of test accuracies and losses for pre-trained and non pre-trained configurations. Subplot (a) captures the test accuracies, while subplot (b) focuses on the test losses. Both metrics are plotted as functions of the epoch count.}
\label{fig:pretraining-nonpretraining}
\end{figure}

The versatility of our DynaLay architecture is further underscored by its robust performance irrespective of whether it undergoes pre-training 
 \cite{HAN2021225}. To elucidate this, Figure \ref{fig:pretraining-nonpretraining} showcases a side-by-side comparison of key performance indicators—test accuracy and test loss—across 100 epochs for both pre-trained and non-pre-trained configurations.

The observed metrics reveal a discernible advantage when employing pre-training. Specifically, the pre-trained model consistently surpasses its non-pre-trained counterpart in both accuracy and loss metrics. This superior performance is attributed to a "warm-up" phase consisting of $20$ epochs of pre-training for the main model. This phase aids in initializing the activations to more optimal states, which likely catalyzes faster and more stable convergence during subsequent training.

\subsection{Role of the Agent in DynaLay's Performance and Adaptability}
Our Agent network also undergoes a pre-training phase lasting the same number of epochs. We preserve the weights \cite{lillicrap2014random} from these pre-training phases and use them as initialization points for the comparative experiment. The findings, delineated in Figure \ref{fig:pretraining-nonpretraining}, validate the efficacy of this approach by showcasing noticeably improved performance metrics for the pre-trained model.

\begin{figure}
  \centering
  \includegraphics[width=\textwidth]{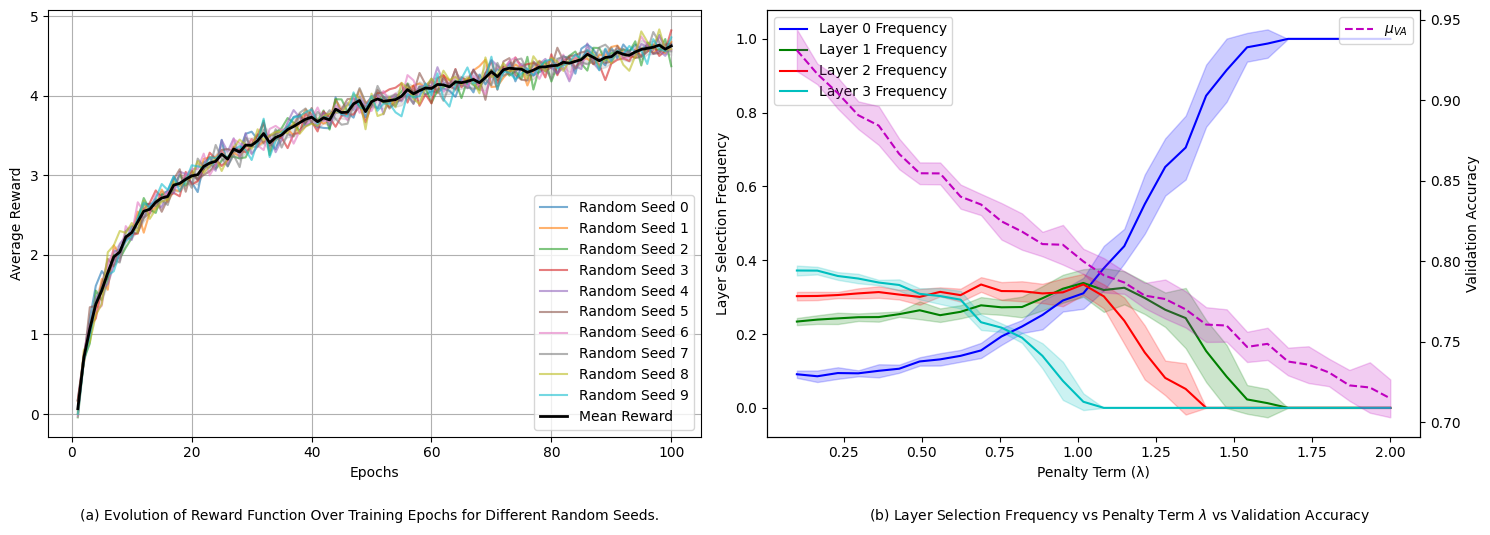}
  \caption{(a) Temporal Dynamics of the Reward Function Across Training Epochs for Distinct Random Seeds. The graph offers an incisive look into the reward function's evolution over the training epochs, integrating both accuracy and computational efficiency into its formulation. Each trace corresponds to a unique random seed, thereby providing a robust measure of the model's resilience to varying initial conditions. (b) Interplay Between Layer Selection Frequency, Penalty Term \( \lambda \), and Validation Accuracy. This tri-axis plot delivers a granular portrayal of how layer selection frequency and validation accuracy respond to changes in the penalty term \( \lambda \).}
  \label{fig:reward-evolution}
\end{figure}

In figure \ref{fig:reward-evolution}: (a) delineates the trajectory of the average reward function throughout the training epochs. A key observation here is the emergent stabilization of the reward function as the training epoch count ascends. This equilibrium is indicative of the agent's escalating competence in judiciously selecting layers that not only enhance performance but also optimize computational expenditure. Further nuance in the agent's decision-making process is captured in Figure \ref{fig:reward-evolution}(b). Each curve on the plot signifies the frequency with which the agent elects to utilize a specific layer across the spectrum of available \( \lambda \) penalty terms. Accompanying these curves is a dashed line that represents the validation accuracy achieved under these conditions. Higher \( \lambda \) values act as deterrents against the selection of computationally burdensome layers, compelling the agent towards more resource-efficient alternatives. This in-depth analysis fortifies the understanding of the agent's role within DynaLay, particularly its aptitude for adaptively managing computational resources without compromising model performance. The agent's proficiency in this balancing act is pivotal for the scalability and applicability of DynaLay across a wide range of tasks and computational settings.

\section{Conclusion and Future Work}
In this study, we have presented DynaLay, a pioneering neural network architecture that employs a reinforcement learning agent for adaptive layer selection, introducing an element of introspection into neural network architectures. Our architecture enables the model to adaptively select and, if necessary, recompute a sequence of layers iterating them until convergence based on the complexity of individual inputs as measured by the model's internal state at the input. Initial results show that DynaLay achieves a compelling balance between computational efficiency and model performance, demonstrating its potential as a versatile solution for deep learning tasks.

As we look towards the future, one immediate avenue for expansion is the incorporation of attention mechanisms within DynaLay. Attention layers have shown remarkable success in various domains and could further augment the model’s adaptability and efficiency. Additionally, we aim to explore our architecture’s capabilities with more intricate models, solidifying its potential as a foundational framework in the field of machine learning.

Our most ambitious goal for future work is to make DynaLay model-agnostic, capable of seamlessly integrating with any existing neural network types—be it CNNs, LSTMs, or Transformer models. This universality would democratize the benefits of dynamic layer selection and computational efficiency across the machine learning landscape. We believe that these future directions will not only validate DynaLay’s efficacy but also position it as a universal architecture, setting new standards for computational adaptability in machine learning models.

\section{Acknowledgment}
This work was supported by  NIH grant 2R01EB006841

\medskip
%\small
%\printbibliography
\bibliographystyle{unsrt}
%\addbibresource{bibliography.bib}
\bibliography{bibliography}

\newpage

\appendix

\section{Agent Architecture}

In DynaLay, The agent comprises a small neural network, often referred to as AgentNet, designed to take as input the current activation state of the main network. It outputs a decision vector that indicates which layer should be activated for the next FPI step.

The agent's input consists of a flattened tensor containing selected features from the main model, the current penalty term $\lambda$, and other relevant metrics. This enables the agent to make informed decisions based on both the data and the current state of the model.

The decision vector output by AgentNet is subjected to a adaptive softmax  \cite{grave2017efficient} operation to convert it into a probability distribution over the available layers. The layer with the highest probability is selected for activation. The agent also employs an epsilon-greedy strategy \cite{dann2022guarantees} during training to explore different layers, thereby preventing the model from getting stuck in local optima.
AgentNet is trained in tandem with the main DynaLay model. The reward function considers both the model's performance in terms of accuracy and the computational efficiency, allowing the agent to learn how to balance these competing objectives effectively.

One of the most compelling features of our agent is its ability to adapt to different tasks and datasets dynamically. This is evident from our experiments with various benchmarks, as discussed in Experiments section.

\section{Backpropagation Algorithm}

The backpropagation algorithm in DynaLay is a cornerstone for optimizing both the main model and the AgentNet. To elucidate the methodological complexity, we first describe the gradient flow within the architecture, followed by the advanced techniques incorporated \cite{liao2018reviving}, 

\subsection{Gradient Flow in DynaLay}
In DynaLay, the gradient flow is bifurcated between the main model and the AgentNet. Given a loss function \( \mathcal{L} \), the gradient \( \nabla_{\text{main}} \mathcal{L} \) for the main model is computed as follows:

\begin{equation}
\nabla_{\text{main}} \mathcal{L} = \frac{\partial \mathcal{L}}{\partial W_{\text{main}}} \quad \text{where \( W_{\text{main}} \) are the weights of the main model.}
\end{equation}

For the AgentNet, the gradient $\nabla_{\text{agent}} \mathcal{L}$ is isolated from affecting the main model's prediction and is computed independently as:

\begin{equation}
\nabla_{\text{agent}} \mathcal{L} = \frac{\partial \mathcal{L}}{\partial W_{\text{agent}}} \quad 
\text{where \( W_{\text{agent}} \) are the weights of the AgentNet.}
\end{equation}

Following the work by \cite{arya2022automatic}, we employ automatic differentiation that caters to the discrete randomness introduced by layer selection. The gradients are computed as:

\begin{equation}
\nabla_{\text{discrete}} \mathcal{L} = \mathbb{E}\left[\frac{\partial \mathcal{L}}{\partial W_{\text{discrete}}}\right]
\end{equation}

We draw from \cite{bolte2022automatic} to handle nonsmooth iterative algorithms, using subdifferentials \( \partial \) to calculate the gradients as:

\begin{equation}
    \nabla_{\text{nonsmooth}} \mathcal{L} = \partial \mathcal{L}(W_{\text{nonsmooth}})
\end{equation}

To improve the scalability, we adapt the asynchronous methods from  \cite{barham2022pathways}. The gradients for each layer \( l \) in an asynchronous setting are calculated as:

\begin{equation}
  \nabla_{l, \text{async}} \mathcal{L} = \nabla_{l} \mathcal{L} + \Delta_{\text{async}}  
\end{equation}

where $ \Delta_{\text{async}} $ is the asynchronous correction term. Echoing the sentiments of \cite{metz2022gradients}, we incorporate auxiliary metrics \( M \) alongside gradients, optimized as:
\begin{equation}
\mathcal{O} = \nabla \mathcal{L} + \alpha M    
\end{equation} 
where \( \alpha \) is a tunable parameter.

\begin{algorithm}
\caption{Advanced Backward Propagation in DynaLay}
\begin{algorithmic}[1]
\Procedure{Backward}{$\text{ctx, grad\_output}$}
    \State \( z_{\star}, \text{layer} \leftarrow \text{ctx.saved\_tensors} \)
    \State \( \text{max\_iter} \leftarrow \text{ctx.max\_iter} \)
    \State \( \text{tol} \leftarrow 1 \times 10^{-5} \)
    \State \( d_z \leftarrow \text{Clone and detach grad\_output} \)
    \State \( I \leftarrow \text{Identity matrix of } d_z \)
    \Comment{Initialize identity matrix}
    \State \( \text{Initialize } \nabla_{\text{agent}} \)
    \Comment{Gradient for the agent network}
    \For{\( k = 1, \ldots, \text{max\_iter} \)}
        \State \( f_z = \text{layer}(z_{\star}) \)
        \State \( J = \frac{\partial f_z}{\partial z_{\star}} \)
        \Comment{Jacobian matrix}
        \State \( \Delta J = I - J \)
        \Comment{Implicit differentiation step}
        \State \( d_{z_{\text{new}}} = d_z \times \Delta J \)
        \Comment{Compute updated gradient}
        \State \( \delta = \frac{\| d_{z_{\text{new}}} - d_z \|}{\| d_{z_{\text{new}}} \|} \)
        \If{\( \delta < \text{tol} \)}
            \State \text{Break}
        \EndIf
        \State \( d_z = d_{z_{\text{new}}} \)
        \State \( \nabla_{\text{agent}} = \text{Advanced Techniques}(\nabla_{\text{agent}}, J, \Delta J) \)
        \Comment{Update agent's gradient}
    \EndFor
    \State \( \text{Update agent network using } \nabla_{\text{agent}} \)
    \Comment{Use optimizer to update agent}
    \State \Return \( d_z, \text{None}, \text{None} \)
\EndProcedure
\end{algorithmic}
\end{algorithm}

\section{Forward Propagation Algorithm}
In addition to the backpropagation algorithm, understanding the forward passis essential for grasping the full scope of DynaLay. The forward process \cite{WANG20131, Greener2023, wang2019satnet} involves solving a fixed-point equation \cite{habiba2021neural} to find a suitable latent state $z$ that serves as the output of the AgentNet.

\begin{algorithm}
\caption{Forward Propagation in DynaLay}
\begin{algorithmic}[1]
\Procedure{Forward}{$\text{ctx}, \text{input\_tensor}, \text{layer}, \text{max\_iter}$}
    \State \( z \leftarrow \text{Clone and detach } \text{input\_tensor} \)
    \State \( z \leftarrow \text{Require gradient for } z \)
    \State \( \text{tol} \leftarrow 1 \times 10^{-2} \)
    \For{\( k = 1, \dots, \text{max\_iter} \)}
        \State \( f_0 \leftarrow \text{Apply layer to } z \)
        \If{\( \text{size of } f_0 \neq \text{size of } z \)}
            \State \( f_0 \leftarrow \text{Interpolate to match dimensions with } z \)
        \EndIf
        \If{\( \frac{\lVert f_0 - z \rVert}{\lVert f_0 \rVert} < \text{tol} \)}
            \State \text{Break}
        \EndIf
        \State \( z \leftarrow f_0 \)
    \EndFor
    \State \Return \( z \)
\EndProcedure
\end{algorithmic}
\end{algorithm}

The algorithm performs a fixed-point iteration to find the state \( z \) that satisfies the equation \( z = f_{\text{layer}}(z) \), where \( f_{\text{layer}} \) is the layer's transformation function. The iteration is performed up to a maximum of \( \text{max\_iter} \) iterations or until the following convergence criterion is met:

\begin{equation}
    \frac{\lVert f_0 - z \rVert}{\lVert f_0 \rVert} < \text{tol}
\end{equation}

where $\text{tol}$ is a predefined tolerance level set to \( 1 \times 10^{-2} \).

As the iterations progress, it is essential to ensure that the dimensions of \( f_0 \) and \( z \) are consistent \cite{almeida1990learning}. 
By employing this forward algorithm, DynaLay achieves a balance between computational efficiency and convergence, ensuring robust and reliable operation across diverse tasks.

\section{Experiments}
\subsection{Impact of Fixed-Point Iteration on Different Model Configurations}
\subsubsection{Objective}
The primary objective of this experiment is to rigorously assess the efficacy of incorporating Fixed-Point Iteration (FPI) into various layers of our \textbf{DynaLay} architecture. We focus on four distinct configurations to perform this assessment:

\begin{enumerate}
    \item \textbf{Model 0}: A straightforward architecture comprised of Layer 1 $\rightarrow$ Layer 2 $\rightarrow$ Layer 3, devoid of FPI.
    \item \textbf{Model 1}: Utilizes FPI exclusively in Layer 1.
    \item \textbf{Model 2}: Employs FPI in both Layer 1 and Layer 2.
    \item \textbf{Model 3}: Applies FPI across all layers (Layer 1 $\rightarrow$ Layer 2 $\rightarrow$ Layer 3).
\end{enumerate}

Our hypothesis posits that the incorporation of FPI into an increasing number of layers will yield a commensurate improvement in key performance metrics, notably test loss and accuracy. Specifically, we project that Model 3 will exhibit superior performance relative to the other configurations, owing to the enhanced complexity and optimization capabilities conferred by FPI.

\begin{figure}
    \centering
    \includegraphics[width=\textwidth]{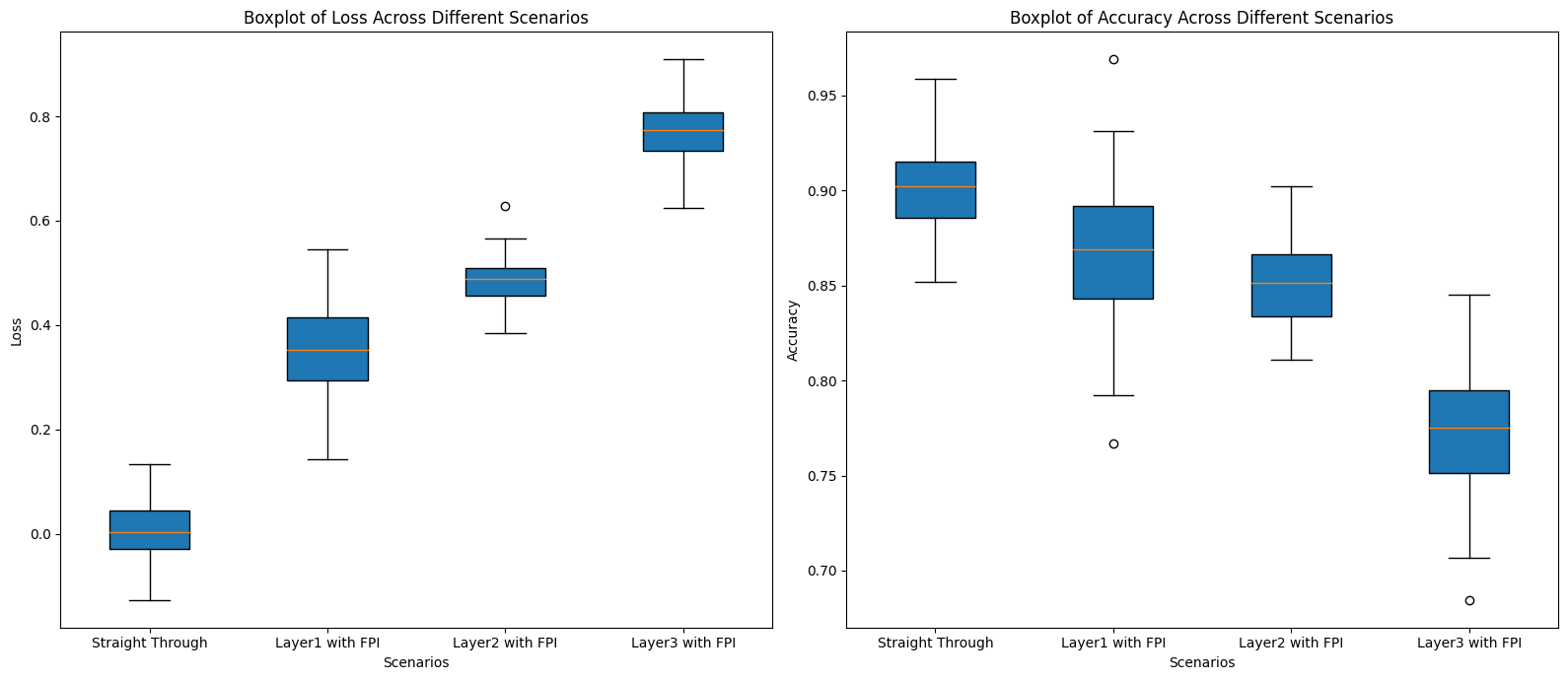}
    \caption{Boxplots illustrating the distribution of test loss and accuracy across four distinct model configurations. The configurations vary in the complexity and number of layers utilizing Fixed-Point Iteration (FPI).}
    \label{fig:impact_of_fpi}
\end{figure}

For evaluation, a test set of 100 samples is employed as seen in Figure \ref{fig:impact_of_fpi}. Each of the four model configurations is subjected to this test set, with both test loss and accuracy being meticulously recorded. It is imperative to note that all models are trained under identical hyperparameter settings to ensure a level playing field.

\subsection{Text-Based Experiments with LSTMs}
To further validate the effectiveness and adaptability of DynaLay, we extend our experiments to text-based tasks employing Long Short-Term Memory (LSTM) networks \cite{memory2010long}. We experiment with two datasets: WikiText-2 for language modeling and IMDB for sentiment analysis.

\subsubsection{Model Configuration}

Our LSTM-based DynaLay model consists of three LSTM layers. The LSTM layers are parameterized as follows:

\begin{equation}
    \begin{aligned}
        f_t &= \sigma(W_f \cdot [h_{t-1}, x_t] + b_f) \\
        i_t &= \sigma(W_i \cdot [h_{t-1}, x_t] + b_i) \\
        o_t &= \sigma(W_o \cdot [h_{t-1}, x_t] + b_o) \\
        \tilde{C_t} &= \tanh(W_C \cdot [h_{t-1}, x_t] + b_C) \\
        C_t &= f_t * C_{t-1} + i_t * \tilde{C_t} \\
        h_t &= o_t * \tanh(C_t)
    \end{aligned}
\end{equation}

In this equation, \( f_t, i_t, o_t \) are the forget, input, and output gates, respectively. \( h_t \) is the hidden state, \( C_t \) is the cell state, and \( x_t \) is the input at time \( t \).

\subsubsection{Experimental Results}

\begin{figure}
    \centering
    \includegraphics[width=\textwidth]{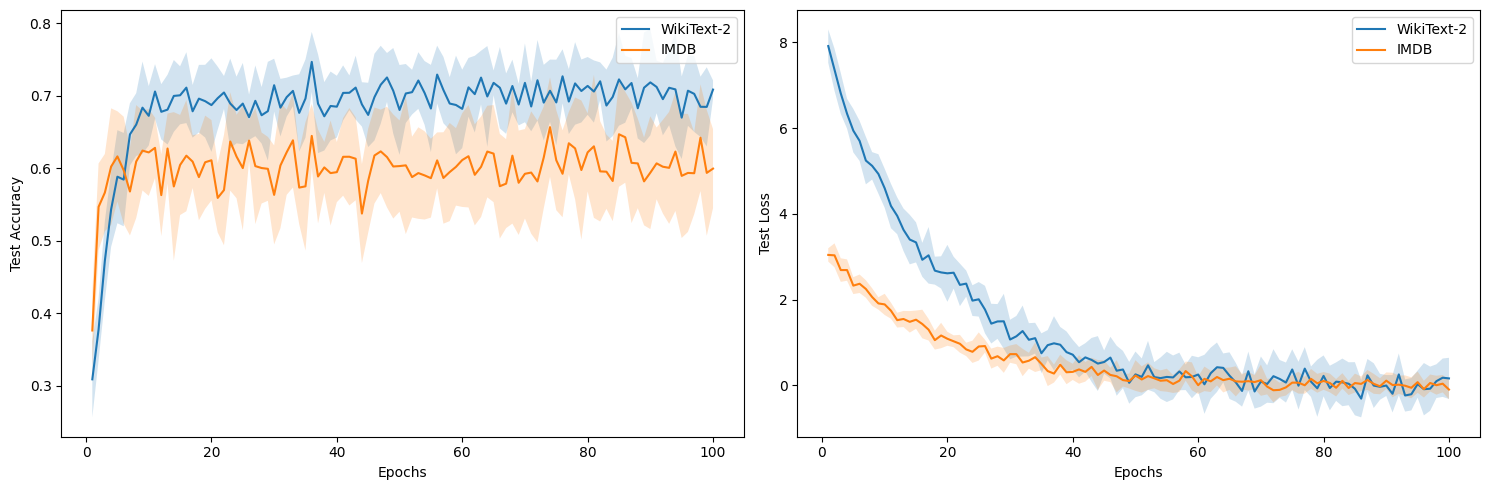}
    \caption{Above plots representing the performance of LSTM-based DynaLay configurations on WikiText-2 and IMDB datasets. Left plot indicates the test accuracy for both datasets and Right plot shows the loss for both datasets}
    \label{fig:lstm_exp}
\end{figure}

We employ Fixed-Point Iteration (FPI) on individual LSTM layers just as in our CNN experiments. The results are summarized in Figure~\ref{fig:lstm_exp}, which shows that the LSTM model with FPI on the third layer achieves the best performance in terms of both accuracy and computational cost.

Through these experiments, we demonstrate that DynaLay's adaptive layer selection mechanism is equally effective for text-based tasks, thereby confirming its versatility across different data modalities and tasks.

\section{Computational Cost and Limitations}

DynaLay's architecture offers significant advantages in terms of computational efficiency and adaptability. However, it is critical to scrutinize the computational cost involved in operating such an intricate architecture, especially when compared to traditional models.
To validate our theoretical findings, we measure the actual computational time taken for training and inference on different datasets as depicted in Table \ref{table:comparison}.

In a typical neural network, the computational steps are often straightforward, involving a series of matrix multiplications and activation functions. In contrast, DynaLay employs a more complex procedure involving Fixed-Point Iteration (FPI) at each layer, in addition to the dynamic layer selection via the Agent mechanism. The computational cost can thus be broken down into the following main components:

\begin{enumerate}
    \item Forward pass through the main model
    \item FPI computation for each dynamically-selected layer
    \item Backward pass involving implicit differentiation
    \item Forward and backward pass through the Agent mechanism
\end{enumerate}

Each of these steps has its own computational overhead \cite{liao2018reviving, banino2021pondernet}, which can grow with the complexity and dimensionality of the data being processed.

\subsection{Limitations}

While DynaLay's architecture offers flexibility and adaptability, it also presents several limitations:

\begin{enumerate}
    \item \textbf{Memory Consumption:} The use of FPI in each layer can substantially increase the memory footprint, especially when dealing with high-dimensional data.
    
    \item \textbf{Convergence:} FPI does not guarantee convergence for all types of layers or activation functions, which might limit its applicability.
    
    \item \textbf{Agent Training:} The Agent mechanism needs to be trained in parallel, adding an extra layer of complexity and computational cost.
    
    \item \textbf{Scalability:} As the architecture grows, so does the computational overhead, potentially making it unsuitable for real-time applications or low-resource environments.
\end{enumerate}

Given these limitations, future work will focus on optimizing the computational steps and exploring techniques for efficient memory management. The goal is to make DynaLay more scalable and applicable to a broader range of machine learning tasks without compromising its computational efficiency.

\end{document}